# Active Learning for Gaussian Process Considering Uncertainties with Application to Shape Control of Composite Fuselage[†]

Xiaowei Yue[1], Yuchen Wen[2], Jeffrey H. Hunt[3], and Jianjun Shi[2]*

[1] Grado Department of Industrial and Systems Engineering, Virginia Polytechnic Institute and State University

[2] H. Milton Stewart School of Industrial and Systems Engineering, Georgia Institute of Technology

[3] The Boeing Company

*(\*Corresponding Author: Dr. Jianjun Shi, jianjun.shi@isye.gatech.edu )*

*Abstract*—In the machine learning domain, active learning is an iterative data selection algorithm for maximizing information acquisition and improving model performance with limited training samples. It is very useful, especially for the industrial applications where training samples are expensive, time-consuming, or difficult to obtain. Existing methods mainly focus on active learning for classification, and a few methods are designed for regression such as linear regression or Gaussian process. Uncertainties from measurement errors and intrinsic input noise inevitably exist in the experimental data, which further affects the modeling performance. The existing active learning methods do not incorporate these uncertainties for Gaussian process. In this paper, we propose two new active learning algorithms for the Gaussian process with uncertainties, which are variance-based weighted active learning algorithm and D-optimal weighted active learning algorithm. Through numerical study, we show that the proposed approach can incorporate the impact from uncertainties, and realize better prediction performance. This approach has been applied to improving the predictive modeling for automatic shape control of composite fuselage.

## I. INTRODUCTION

Active learning is a type of iterative supervised learning which focuses on maximizing information acquisition with limited samples. In statistics literature, this process is also called optimal experimental design, or sequential design. The main idea of active learning is to iteratively pose "query" or "design" to explore the most informative new experimental samples according to the information obtained from the current samples.

In many machine learning applications, especially in some industrial systems, the explanatory data $X$ are rich and easy to get, but the response data $Y$ are very expensive, time-consuming, or difficult to obtain. For example, when training autonomous driving algorithms, a lot of media (e.g., images, videos) require that oracle users mark them with particular labels, such as "vehicle", "street sign" or "road lines". It can be tedious, redundant and time-consuming to annotate lots of these instances. In composite fuselage shape control problem, it is very expensive, and time-consuming to collect the dimensional shape under specific actuators' forces. Other examples such as speech recognition, factual information extraction, computational biology, etc. can be found [1]. In these scenarios, active learning is well-motivated because it can reduce the samples needed as well as obtain sufficient information for parameter inference of predictive modeling.

Different sampling strategies have been proposed to realize active learning. In machine learning literature, these strategies are classified into query synthesis, stream-based selective sampling, and pool-based sampling [1]. In statistics literature, space-filling designs, criterion-based designs, and expected improvement (EI) algorithm have been proposed for sequential design. For the comparison between active learning and sequential design, they have several common characteristics. For example, (i) they both iteratively pose "query" or "design" to explore the most informative new samples according to the information obtained from the current samples; (ii) some criteria are similar for active learning and sequential design. The uncertainty sampling in active learning and sequential designs are based on certain optimality criteria (e.g., mean squared prediction error and maximum entropy) that are mathematically equivalent [1, 2].

However, there exist some differences between active learning and sequential design, in particular, (i) sequential design literature is mainly for regression and prediction problems. While active learning literature mainly focuses on classification problems in machine learning, and a few papers focus on regression with extrinsic uncertainties but not incorporating impacts from intrinsic input uncertainties; (ii) In sequential design, usually the experiment will be conducted at the selected points from an input space; while in active learning, the experiment sampling may be from a large pool of existent unlabeled data, and the learner can conduct the experiment at a selected point or discard it. In Section 2, we will review the literature from two perspectives: machine learning (computational learning) and statistics (sequential design), in detail. Other than Section 2, we use active learning as a consistent terminology.

In this paper, we develop two active learning algorithms for Gaussian process model with uncertainties. Uncertainties inevitably exist in the input and output data of any system. From the predictive modeling perspective, several approaches have been developed to incorporate input and output uncertainties. Ankenman et al. separated the uncertainties into intrinsic uncertainty inherent to stochastic simulation, and extrinsic uncertainty about unknown response surface [3]. They proposed the stochastic Kriging model for stochastic computer experiments. Cervone and Pillai investigated Gaussian process regression with input uncertainty from measurement errors and showed that approximate methods for incorporating location measurement error are essential to valid parameter inference [4]. Wang et al. compared the best linear unbiased predictor and stochastic Kriging predictor, and proved their asymptotic properties [5]. For the sampling perspective, the existing active learning methods cannot be straightforwardly extended to Gaussian process with uncertainties. In this paper, our main contribution is to propose active learning algorithms for Gaussian process regression considering both intrinsic and extrinsic uncertainties. Specifically, the proposed active learning algorithms could maximize acquisition of various information (for instance, actuators' uncertainty, composite parts variability, modeling nonlinearity, and measurement errors).

The proposed active learning approach is applied to improving the predictive modeling for the automatic shape control system of composite fuselage. The automatic shape control system is essential to reducing the dimensional deviations of composite fuselage as well as the flow time of the aircraft assembly process. Ten actuators are installed cross the edge of the composite fuselage. These actuators are able to provide push and pull forces to change the in-plane shape of the composite fuselage. The automatic shape control system effectively and efficiently adjusts composite fuselages to the optimal configuration [6]. Precise prediction of the dimensional shape of composite fuselage is essential for the shape control, while it is challenging to develop predictive modeling because of multiple uncertainties. A surrogate model considering part uncertainty, actuator uncertainty, and model uncertainty was developed to link the relationship between actuators' forces and dimensional deviations [7]. This approach assumes that the training samples are sufficiently large, and it does not consider efficient experimental sampling and effective information acquisition. However, obtaining the experimental samples in the composite fuselage assembly process is time-consuming and expensive. To find an approach to minimize the sample size and improve predictive modeling in a sequential way would be beneficial. In this paper, we propose active learning algorithms for maximizing information extractions for Gaussian process model considering uncertainties. We firstly derive the log-likelihood for the general Gaussian process model considering uncertainties, including stochastic model and surrogate model considering uncertainties. Then we take two kinds of information measure into consideration: variance-based information measure, and Fisher information measure. Based on the information measure, we proposed the variance-based weighted active learning algorithm and D-optimal weighted active learning algorithm. The initial design and stopping criterion have also been explored. The proposed active learning strategies have been applied to improving the predictive modeling of composite fuselage shape control.

The remainder of this paper is organized as follows: Section II illustrates the literature review about active learning and sequential design from machine learning perspective and statistics perspective. Section III describes the predictive modeling based on Gaussian process, and then proposes the active learning approach for Gaussian process model considering uncertainties. Section IV presents the case study, including an introduction for automatic shape control of composite fuselage, validation procedure, evaluation criteria, and comparison between the performance of our proposed method and benchmark methods. Finally, a brief summary is provided in Section V.

## II. Literature Review

In this section, we review the existing active learning and sequential design methodologies from machine learning (computational learning) and statistics (sequential design) perspectives.

### A. Literature from Machine Learning Domain

In computational learning theory literature, active learning is sometimes called "query learning". There are three kinds of scenarios: query synthesis, stream-based selective sampling, and pool-based sampling [1]. In query synthesis, the learner may request any unlabeled data instance in the input space. While in stream-based selective sampling, each unlabeled instance is typically drawn one at a time from the input space and then the learner decides whether to query or discard it. For pool-based sampling, there are a set of labeled data and a large pool of unlabeled data available, and queries are selected from the pool. Detailed descriptions about these three scenarios can be found in [1].

Settles and Olsson provided a very detailed literature review on active learning for classification problems [1, 8]. While in this paper, we mainly focus on the literature review on active learning for regression problems.

Lewis and Catlett proposed one active learning strategy called uncertainty sampling [9]. In this strategy, the learner regards the unlabeled samples with the largest uncertainty as the most informative samples. Therefore, the learner queries these unlabeled samples in order to generalize information from the current samples. In regression, the learner can query the unlabeled samples for which the model has the highest prediction variance. Under the Gaussian assumption, this prediction variance-based uncertainty sampling approach is equivalent to the entropy-based uncertainty sampling because the entropy is a monotonic function of its variance [1]. Other than uncertainty based criteria, some algorithms introduced diversity constraints to avoid selecting similar samples for model training. For example, Wang et al. proposed an active learning algorithm to integrate uncertainty and diversity via sparse modeling in the sample selection [10][11]. Yang et al. also proposed an uncertainty sampling with diversity maximization to improve multi-class active learning [12]. Cohn Ghahramani and Jordan proposed an active learning strategy that selects design points by minimizing the integrated average variance of the learner [13]. Sugiyama proposed a new active learning method called *ALICE* (Active Learning using the Importance-weighted least-squares learning based on Conditional Expectation of the generalization error) [14]. This method predicts conditional expectation of the generalization error given training input points, while most of the existing methods predict the full expectation of the generalization error. Burbidge et al. investigated an active learning strategy for regression based on Query by Committee [15], which considers choosing sequential points according to the average expected variance over the reference points. Sugiyama and Rubens developed a new ensemble active learning approach for solving active learning and model selection in linear regression simultaneously [16]. However, this approach is mainly designed for linear regression, and not easy to extend to other models like the Gaussian process. Pasolli and Melgani proposed two active learning strategies for Gaussian Process (GP) regression [17]. One is based on adding samples that have large kernel distance from the current training samples, which considers space filling properties. The other exploits an intrinsic GP regression outcome to pick up the samples with the largest variance. Cai et al. proposed a new active learning framework for regression called Expected Model Change Maximization (EMCM) [18], which aims to choose the examples that lead to the largest change to the current model. However, the EMCM is sensitive to outliers, which may result in non-stationary parameter estimations. Schreiter et al. proposed a safe exploration for active learning with Gaussian processes [19]. A differential entropy criterion was used to explore the relevant data regions. These existing active learning methods do not straightforwardly extend to incorporate uncertainties in the automatic shape control system of composite fuselages.

B. *Literature from Statistics Domain*

In the statistics domain, the sequential design is to propose experiment designs at a limited number of times, and inputs/responses from the previous design may impact the following design. The basic idea behind the sequential design is to select input points that will allow us to model and minimize the discrepancy between the output from the computer model and predictions from the surrogate model.

For experimental designs relevant to computer experiments, there are two categories: space-filling designs and criterion-based designs. Space-filling designs (e.g. Latin hypercube designs, maximin designs, Sobol's sequence [20]) assume that samples provide information equally across the entire input space, which encourages the exploration among the whole input space. However, these designs are not adaptive to the information from the response surface. Designs based on certain optimality criteria, such as mean squared prediction error [21] and entropy [22], make full use of information from both inputs and outputs. However, they do not consider the shape of the response surface.

Expected improvement (EI) algorithm is a global optimization algorithm proposed by Mockus [23] and then brought to the field of computer experiments [24]. The main idea of EI algorithm is to identify the nature of input-output relationships, and then subsequently choose design points one at a time, or in groups, to maximize the expected improvement on the objective. Williams et al. modified the EI algorithm by considering both control and environmental variables [25], which computed the posterior expected improvement over the current optimum for each untested point, and then selected the next point to maximize modified EI. They further extended the EI algorithm to a bivariate modified expected improvement algorithm, which realizes sequential design for computer experiments where there is a bivariate response [26]. Vazquez and Bect investigated the convergence properties of the EI algorithm [27]. Provided that the objective function belongs to the reproducing kernel Hilbert space, the EI algorithm produces a dense sequence of evaluation points in the search domain.

In addition, Lam proposed a modified integrated mean squared prediction error (IMSPE) criterion by imposing a penalty to prevent the additional design points from clustering together [2]. This method can realize the trade-off between exploration and exploitation. Deng et al. pointed out that there were two kinds of approaches to generate sequential designs: stochastic approximation and optimal design [28]. The optimal design approach has better performance when the assumed model is the true model, but it is not robust to model assumption. They used a combination of stochastic approximation and D-optimal designs to judiciously select the design points [28]. By maximizing the estimated variance of the hyperplane, they placed the next point at the location of the greatest uncertainty. The proposed method improved the process of money laundering detection. Crombecq et al. proposed a hybrid sequential design strategy which used a Monte-Carlo-based approximation of a Voronoi tessellation for exploration and local linear approximations of the simulator for exploitation [29]. The advantage of this method is that it is independent of the model type, and can be used in heterogeneous modeling environments.

From an engineering perspective, there are several literature presenting the sequential strategies for other objectives, such as measurement and detection. Jin et al. proposed a sequential measurement strategy for efficient wafer geometric profile estimation [30]. This strategy reduced the number of samples measured in wafers as well as provided an adequate accuracy for quality feature estimation. The sequential samples are chosen based on the gradient and error of profile prediction. Hao proposed a sequential sampling strategy called Adaptive Kernelized Maximum-Minimum Distance (AKM2D) to speed up inspection and anomaly detection process [31]. The proposed method realized the trade-off between space filling sampling (exploration) and focused sampling near the anomalous region (exploitation). Xiang et al. developed novel distribution-free control chart via cucconi statistic for joint monitoring of location and scale parameters with extremely small samples [32]. However, these methods are mainly focused on efficient detection, not for predictive modeling.

### III. ACTIVE LEARNING FOR GAUSSIAN PROCESS CONSIDERING UNCERTAINTIES

Gaussian process models have been widely used as surrogate models of expensive deterministic computer simulations. It has many advantages, such as good prediction performance, good flexibility from multiple correlation choices, complete mathematical properties in both Bayesian and frequentist statistical framework, and capability of uncertainty quantification. In Section II, several active learning strategies relevant to the Gaussian process models have been reviewed, including uncertainty sampling [9, 17], entropy-based active learning [19], and EI algorithms [24-26]. These strategies work well for the data acquisition of the deterministic computer simulations, while in some other cases, there exist multiple input or output uncertainties in the datasets. To the best of our knowledge, there lacks a tailored active learning strategy for Gaussian process considering uncertainties. In this paper, we first review two

Gaussian process models considering uncertainties for automatic shape control of composite fuselage. One is the stochastic Kriging, and the other is the surrogate model considering uncertainties. These models can provide a very accurate prediction for composite fuselage shape control. Then, we proposed new active learning strategies for Gaussian process considering uncertainties.

A. *Gaussian Process with Nugget Effects: Stochastic Kriging*

Cressie represented measurement error or uncertainties as a nugget effect for Gaussian process [33]. Ankenman et al. integrated the intrinsic uncertainty and extrinsic uncertainty with their stochastic Kriging model [3]. Cervone and Pillai investigated Gaussian process regression with input uncertainty from measurement errors and showed that approximate methods for incorporating location measurement error are essential to valid parameter inference [4]. In this section, we summarize the Gaussian process model with nugget effects for composite fuselage shape control.

Consider a set of design setting $\mathcal{X}$, which includes pairs $\{\boldsymbol{F}_t, n_t\}$, $t = 1,2,\ldots,k$, where $\boldsymbol{F}_t$ is the input vector (actuators' forces) with dimension of $1 \times q$, $n_t$ is the number of replications for the design point $\boldsymbol{F}_t$. Consider that the p-dimension output (dimensional deviations) for the design point $\boldsymbol{F}_t$ is $\boldsymbol{Y}(\boldsymbol{F}_t)$, where each element $Y_{ij}(\boldsymbol{F}_t)$ denotes the $j^{\text{th}}$ variable of the output vector under the $i^{\text{th}}$ replication at the input $\boldsymbol{F}_t$. We use the set $\mathcal{D}$ to represent the input/output pairs $\{\boldsymbol{F}_t, \boldsymbol{Y}(\boldsymbol{F}_t)\}$. Considering both a noise term $\boldsymbol{\varepsilon}_i$ and a stochastic process term $z_j(\boldsymbol{F})$ that are regarded as intrinsic uncertainty (measurement error) and extrinsic uncertainty respectively, the Gaussian process with nuggets effects (or called the stochastic Kriging) model can be developed as

$$Y_{ij}(\boldsymbol{F}) = \boldsymbol{F}\boldsymbol{S}_j + z_j(\boldsymbol{F}) + \varepsilon_{ij}, \tag{1}$$

where $\boldsymbol{S}_j$ represents the sensitivity matrix corresponding to the $j^{\text{th}}$ response, $j = 1,2,\ldots,p$. $z_j(\boldsymbol{F})$ is a stochastic process that represents the extrinsic uncertainty relevant to functional mapping. Specifically, the stochastic process $z_j(\boldsymbol{F})$ is assumed to be a Gaussian process $\boldsymbol{GP}(\boldsymbol{0}, \boldsymbol{\Sigma}_{zj})$ with covariance matrix $\boldsymbol{\Sigma}_{zj}$ with dimension $k \times k$. For any two vectors of the actuators' forces, the covariance is $\boldsymbol{\Sigma}_{zj}(\boldsymbol{F}_m, \boldsymbol{F}_n) = cov[z_j(\boldsymbol{F}_m), z_j(\boldsymbol{F}_n)]$. Let $\boldsymbol{\Sigma}_{zj}(\boldsymbol{F}_0,\cdot) = (cov[z_j(\boldsymbol{F}_0), z_j(\boldsymbol{F}_1)], \ldots, cov[z_j(\boldsymbol{F}_0), z_j(\boldsymbol{F}_k)])$ as the covariance between $z_j$'s at design points and new actuators' forces $\boldsymbol{F}_0$. The intrinsic noise $\varepsilon_{ij}$ is assumed to be independent and identically distributed with a normal distribution $\varepsilon_{ij} \sim \mathbb{N}(0, \sigma_{\varepsilon j})$. the noise covariance matrix is $\boldsymbol{\Sigma}_{\varepsilon j}$, a $k \times k$ covariance matrix with $(a,b)$ element is $\text{Cov}[\sum_{i=1}^{n_a} \varepsilon_{ij}(\boldsymbol{F}_a)/n_a, \sum_{i=1}^{n_b} \varepsilon_{ij}(\boldsymbol{F}_b)/n_b]$. The sample mean at $\boldsymbol{F}_t$ as $\overline{Y}_j(\boldsymbol{F}_t) = \sum_{i=1}^{n_t} Y_{ij}(\boldsymbol{F}_t)/n_t$. Let $\overline{\boldsymbol{Y}}_j = (\overline{Y}_j(\boldsymbol{F}_1), \ldots, \overline{Y}_j(\boldsymbol{F}_k))^T$, and Let $\boldsymbol{F}_{DOE} = (\boldsymbol{F}_1; \ldots; \boldsymbol{F}_k)$ denote all the design points of the actuators' forces. Let $\boldsymbol{R}_j(\boldsymbol{F}_0,\cdot) = (\text{Cov}[Y_j(\boldsymbol{F}_0), \overline{Y}_j(\boldsymbol{F}_1)]; \ldots; \text{Cov}[Y_j(\boldsymbol{F}_0), \overline{Y}_j(\boldsymbol{F}_k)])$. The covariance between vector $\boldsymbol{a}$ and vector $\boldsymbol{b}$ satisfies $\boldsymbol{\Sigma}_{zj}(\tau_j^2, \boldsymbol{\theta}, \boldsymbol{a}, \boldsymbol{b}) = \tau_j^2 R_{zj}(\boldsymbol{\theta}, \boldsymbol{a} - \boldsymbol{b})$, where $R_{zj}(\boldsymbol{\theta}, \boldsymbol{a} - \boldsymbol{b})$ is one of the correlation functions. With spatial correlation, $z_j(\boldsymbol{F}_m)$ and $z_j(\boldsymbol{F}_n)$ tend to be similar (e.g. $R_{zj,mn} = R_{zj}(\boldsymbol{\theta}, \boldsymbol{F}_m, \boldsymbol{F}_n)$ tends to be large) if $\boldsymbol{F}_m$ and $\boldsymbol{F}_n$ are close.

According to the literature [3, 4, 7], we know the best MSPE (mean square prediction error) linear unbiased predictor as

$$\hat{Y}_j(\boldsymbol{F_0} \mid \mathcal{D}) = \boldsymbol{F_0}\boldsymbol{S}_j + \boldsymbol{R}_j(\boldsymbol{F_0},\cdot)^T[\boldsymbol{\Sigma}_{zj} + \boldsymbol{\Sigma}_{\varepsilon j}]^{-1}(\overline{\boldsymbol{Y}}_j - \boldsymbol{F}_{DOE}\boldsymbol{S}_j) \tag{2}$$

B. *Surrogate Model considering Uncertainties*

The stochastic Kriging model uses a nugget effect to approximate both input uncertainties and output uncertainties. Yue et al. proposed a surrogate model considering part uncertainty, actuator uncertainty, and model uncertainty, which is a Gaussian process model with consideration of uncertainties in detail [7]. A surrogate model considering uncertainties was proposed as

$$Y_{ij}(F_t) = F_t S_j + F_t \tilde{S}_j + \tilde{F}_t S_j + z_j(F_t) + \varepsilon_{ij}(F_t), \quad (3)$$

where $i = 1,2,\ldots,n_t$; $j = 1,2,\ldots,p$; $p$ is the number of output responses (key dimensional features). $F_t$ is the target actuators' forces vector; $\tilde{F}_t$ is a vector of an additional random deviation of actuators' forces that results from the actuators' uncertainty with distribution $\mathbb{N}(0, \Sigma_F)$. It can be obtained from the tolerance of actuators instruction; $F_t + \tilde{F}_t$ represents the true actuators' force vector. $S_j$ is an ideal sensitivity vector (column vector) and $\tilde{S}_j$ represents the random sensitivity vector variability from the part uncertainty, which is assumed to follow $\mathbb{N}(0, \Sigma_S)$. Both $S_j$ and $\Sigma_S$ are unknown. $z_j(F_t)$ is assumed to be a stationary Gaussian process $z_j \sim GP(0, \Sigma_{zj})$. $\varepsilon_{ij}(F_t)$ is assumed to follow an independent normal distribution $\varepsilon_{ij}(F_t) \sim \mathbb{N}(0, \sigma^2_{\varepsilon j}(F_t))$, which represents the inherent simulation variability in a stochastic simulation, or measurement errors in a physical experiment. $\varepsilon_{ij}$, $\tilde{F}_t$, $\tilde{S}_j$, and $z_j$ are assumed to be mutually independent, and their higher order interaction term $\tilde{F}_t \cdot \tilde{S}_j$ is assumed to be zero. The model can be interpreted as a decomposition of the response $Y_{ij}(F_t)$ into three parts: a regression term $F_t S_j + F_t \tilde{S}_j + \tilde{F}_t S_j$, a Gaussian process term $z_j(F_t)$, and a noise term $\varepsilon_{ij}(F_t)$.

Assume that $S_j$, $\Sigma_{zj}$, $\Sigma_F$, $\Sigma_\varepsilon$, and $\Sigma_S$ are known, the best MSPE (mean square prediction error) linear unbiased predictor can be derived as

$$\hat{Y}_j(F_0 \mid \mathcal{D}) = F_0 S_j + R_j^T(F_0, \cdot) R_j^{-1}(\bar{Y}_j - F_{DOE} S_j), \quad (4)$$

where $R_j = F_{DOE} \Sigma_S F_{DOE}^T + \Sigma_{zj} + S_j^T \Sigma_F S_j \cdot I + \Sigma_{\varepsilon j}$, and $R_j(F_0, \cdot) = \left(F_0 \Sigma_S F_1^T + \tau_j^2 R_{zj}(\theta, F_0 - F_1); \ldots; F_0 \Sigma_S F_k^T + \tau_j^2 R_{zj}(\theta, F_0 - F_k)\right)$. The best MSPE linear unbiased predictor is also simply called a best linear unbiased predictor (BLUP). More detail related to the algorithm for the surrogate model considering uncertainties can be found in [7].

The surrogate model considering uncertainties in Equation (4) analyzes the different sources of uncertainties in the composite fuselage shape control system, while the stochastic Kriging predictor in Equation (2) approximates all the uncertainties by introducing a nugget effect. Wang et al. proved that the stochastic Kriging and Gaussian process with input location errors asymptotically converge to the same limit [5]. In our active learning strategy design, we will analyze both cases.

C. *Information Measure*

Active learning is an iterative data selection algorithm for maximizing information acquisition and improving model performance with limited training samples. Firstly, we need to propose the information measure for Gaussian process considering uncertainties.

Suppose $\mathfrak{R}_j(F_0,\cdot)$ and $\mathfrak{R}_j$ describe the covariance between $F_0$ and historical samples and the covariance among historical samples for general Gaussian process models considering uncertainties. When the model is the surrogate model considering uncertainties, $\mathfrak{R}_j(F_0,\cdot) = \left( F_0 \Sigma_S F_1^T + \tau_j^2 R_{zj}(\boldsymbol{\theta}, F_0 - F_1); \ldots; F_0 \Sigma_S F_k^T + \tau_j^2 R_{zj}(\boldsymbol{\theta}, F_0 - F_k) \right)$, $\mathfrak{R}_j = F_{DOE} \Sigma_S F_{DOE}^T + \Sigma_{zj} + S_j^T \Sigma_F S_j \cdot I + \Sigma_{\varepsilon j}$; When the model is the stochastic Kriging model, $\mathfrak{R}_j(F_0,\cdot) = \left( \tau_j^2 R_{zj}(\boldsymbol{\theta}, F_0 - F_1); \ldots; \tau_j^2 R_{zj}(\boldsymbol{\theta}, F_0 - F_k) \right)$, $\mathfrak{R}_j = \Sigma_{zj} + \Sigma_{\varepsilon j}$. Suppose $\Sigma_{\varepsilon j} = \sigma_j^2 I$, $\Sigma_S = \varphi_j^2 I$. Let $\Theta$ represent the key parameter set, for example in the stochastic Kriging model, $\Theta = \{\tau_j^2, \boldsymbol{\theta}_j, \sigma_j^2\}$, while in the surrogate model considering uncertainties, $\Theta = \{\tau_j^2, \boldsymbol{\theta}_j, \sigma_j^2, \varphi_j^2\}$.

Under the multivariate normal distribution, the log-likelihood function of $(S_j, \tau_j^2, \boldsymbol{\theta}, \Sigma_S)$ is

$$\mathcal{L}(S_j, \Theta \mid \mathcal{D}) = -\frac{1}{2}\ln[(2\pi)^k] - \frac{1}{2}\ln[\det(\mathfrak{R}_j)] - \frac{1}{2}(\overline{Y}_j - F_{DOE} S_j)^T \mathfrak{R}_j^{-1} (\overline{Y}_j - F_{DOE} S_j) \quad (5)$$

We can derive the estimated parameter $\widehat{S}_j$ by making the first derivative of $\mathcal{L}(S_j, \tau_j^2, \boldsymbol{\theta}_j, \Sigma_S)$ to $S_j$ be equal to zero. Then we can get the generalized least-square estimation of the parameter $\widehat{S}_j$

$$\frac{\partial \mathcal{L}}{\partial S_j} = (\overline{Y}_j - F_{DOE} S_j)^T \mathfrak{R}_j^{-1} F_{DOE} = 0$$
$$\widehat{S}_j(\Theta) = (F_{DOE}^T \mathfrak{R}_j^{-1} F_{DOE})^{-1} F_{DOE}^T \mathfrak{R}_j^{-1} F_{DOE} \overline{Y}_j \quad (6)$$

One straightforward and widely used measure is variance. The variance of the predictor $\widehat{Y}_j(F_0 \mid \mathcal{D})$ is $\mathrm{Var}_j(F_0 \mid \mathcal{D}) = \tau_j^2 - \mathfrak{R}_j^T(F_0,\cdot) \mathfrak{R}_j^{-1} \mathfrak{R}_j(F_0,\cdot)$.

The other information measure is the Fisher information matrix $I(\Theta) \in \mathbb{R}^{(m+2)\times(m+2)}$, which is calculated by

$$I_{ab}(\Theta \mid \mathcal{D}) = -\mathbb{E}\left[ \frac{\partial^2}{\partial \Theta_a \partial \Theta_b} \mathcal{L}(\Theta) \mid \Theta \right] \quad (7)$$

where $a, b = 1, \cdots, m+2$.

Based on Equation (6) and Equation (7), the Fisher information matrix of $\widehat{\Theta}$ is

$$I_{ab}(\widehat{\Theta} \mid \mathcal{D}) = \left[\frac{\partial S_j}{\partial \Theta_a}\right]^T F_{DOE}^T \mathfrak{R}_j^{-1} F_{DOE} \left[\frac{\partial S_j}{\partial \Theta_b}\right] + \frac{1}{2} Tr\left[ \mathfrak{R}_j^{-1} \frac{\partial \mathfrak{R}_j}{\partial \Theta_a} \mathfrak{R}_j^{-1} \frac{\partial \mathfrak{R}_j}{\partial \Theta_b} \right] \quad (8)$$

where $\mathfrak{R}_j$ is the covariance matrix of the Gaussian process; and the first-order derivative of the coefficients $S_j$ with respect to each entry in the parameter vector can be represented as

$$\frac{\partial S_j}{\partial \Theta_a}(\widehat{\Theta})$$
$$= -\left(F_{DOE}^T \mathfrak{R}_j^{-1} F_{DOE}\right)^{-1} \left[ F_{DOE}^T \mathfrak{R}_j^{-1} \frac{\partial \mathfrak{R}_j}{\partial \Theta_a} \mathfrak{R}_j^{-1} (\overline{Y}_j \quad (9) - F_{DOE} S_j) \right]$$

According to Cramer-Rao inequality explanation for fisher information matrix [34], the inverse of Fisher information sets a lower bound on the variance of the model's parameter estimates. Maximizing the Fisher information is equivalent to minimizing the lower bound on the variance of parameter estimations. When the Fisher information is a matrix, we minimize the determinant of the inverse information matrix. This is called D-optimality in the optimal experimental design.

D. *Active Learning*

In this section, we propose algorithms to select the next sample sequentially, which is the main implementation of active learning. Suppose the next samples can be selected from a pool of candidates. We denote these candidate samples by $\mathcal{F} = \{\widetilde{F}_1, \widetilde{F}_2, \cdots, \widetilde{F}_N\}$. $N$ denotes the size of the pool $\mathcal{F}$. It is worth mentioning that the choice of candidate pool $\mathcal{F}$ is very important. The best active learning strategy should perform a trade-off between exploitation and exploration, where the exploitation suggests selecting samples in regions which were previously identified to be interesting. On the other hand, the exploration involves selecting samples in unrepresented regions of the design space. In our active learning strategy, the exploitation is relevant to algorithms of selecting next samples from the candidate pool, while the candidate pool determines the exploration. In our algorithms, the maximin Latin Hypercube Design [35], which demonstrates good space-filling properties and first-dimension projection properties, is implemented to obtain the samples in the candidate pool.

There are two information measures according to Section III.C, the output variance of its prediction, and the Fisher information matrix. Firstly, we develop a variance-based weighted active learning (VWAL) algorithm for Gaussian process considering uncertainties. That means the next sample is selected based on

$$F_{new} = \underset{F \in \{\widetilde{F}_1, \widetilde{F}_2, \cdots, \widetilde{F}_N\}}{\arg\max} \sum_{j=1}^{p} W_j \cdot \text{Var}_j(F \mid \mathcal{D}) = \underset{F \in \{\widetilde{F}_1, \widetilde{F}_2, \cdots, \widetilde{F}_N\}}{\arg\max} \sum_{j=1}^{p} \tau_j^2 - \mathfrak{R}_j^T(F_0, \cdot) \mathfrak{R}_j^{-1} \mathfrak{R}_j(F_0, \cdot) \quad (10)$$

where $F_{new}$ is the next sample to be queried, $\text{Var}_j(F \mid \mathcal{D})$ is the variance of the predictor $\widehat{Y}_j(F_0 \mid \mathcal{D})$ at the $j^{\text{th}}$ critical dimension of the composite fuselage, and $W_j$ is weight coefficient for the $j^{\text{th}}$ critical dimension and we suppose $\sum_j W_j = 1$. The $W_j$ ($j = 1,2, \ldots, p$) is determined by the engineering-domain knowledge. Here $W_j$ describes the relative degree of importance of critical dimensional feature $j$. For instance, the lower part of the fuselage may have higher weights under some scenarios. Under Gaussian assumption, the entropy of a random variable is a monotonic function of its variance.

For the implementation of the active learning algorithm, firstly, we estimate model parameters for the Gaussian process model considering uncertainties by maximizing the log-likelihood function (5). Then we calculate the variance of predictors for each sample in the candidate pool. Next, the new sample point $F_{new}$ can be selected based on solving Equation (10). The experiment will be conducted to collect the oracle response $Y(F_{new})$. The existing sample set $\mathcal{D}$ will be updated by adding the new sample point and corresponding response. The parameters will be iteratively updated and the new samples will be selected actively, until the iteration number reaches the maximum iteration $N_{iter}$ or the error of the model $E_{GP}$ is smaller than a specific threshold. The pseudo code of this active learning algorithm is summarized in Algorithm 1.

---

**Algorithm 1**: Variance-based Weighted Active Learning (VWAL) for Gaussian Process Considering Uncertainties

---

Require: $\mathcal{D}, W_j$

1: $i = 1$

2: Estimate parameters by maximizing log-likelihood function (5) on $\mathcal{D}$

3: Calculate the variance of the predictor $\widehat{Y}_j(F_0 | \mathcal{D}), (j = 1, 2, \ldots, p)$ for each sample

4: **while** $(i \leq N_{iter})$ && $(E_{GP} > \text{threshold})$ **do**

5:      get $F_{new}$ from solving Equation (10)

6:      Implement experiment for $F_{new}$ and obtain $Y(F_{new})$

7:      $\mathcal{D} \leftarrow \{\mathcal{D} \cup (F_{new}, Y(F_{new}))\}$

8:      Estimate parameters by maximizing log-likelihood function (5) on $\mathcal{D}$

9:      Calculate the variance of the predictor $\widehat{Y}_j(F_0 | \mathcal{D}), (j = 1, 2, \ldots, p)$ for each sample

10:     $i = i + 1$

11: **end while**

---

Based on another information measure, the Fisher information matrix, we also develop a D-optimal weighted active learning (DOWAL) algorithm for Gaussian process considering uncertainties. That means the next sample is selected based on

$$F_{new} = \underset{F \in \{\tilde{F}_1, \tilde{F}_2, \cdots, \tilde{F}_N\}}{\arg\min} \sum_{j=1}^{p} W_j \cdot \det\left(\left[I_{ab}(\widehat{\Theta} | \mathcal{D}, F)\right]^{-1}\right) \tag{11}$$

The inverse of Fisher information sets a lower bound on the variance of the model's parameter estimates, which is known as the Cramér-Rao inequality. By minimizing the determinant of the inverse Fisher information matrix (D-optimality), we can maximize the information acquisition in each step of active learning process. Furthermore, the D-optimality is relevant to minimizing the differential posterior entropy of the parameter estimation [36].

Similarly, we can summarize the pseudo code of this D-optimal weighted active learning algorithm for Gaussian process model considering uncertainties in Algorithm 2.

---

**Algorithm 2**: D-Optimal Weighted Active Learning (DOWAL) for Gaussian Process Considering Uncertainties

---

Require: $\mathcal{D}, W_j$

1: $i = 1$

2: Estimate parameters by maximizing log-likelihood function (5) on $\mathcal{D}$

3: Calculate the Fisher information matrix by solving Equations (8) and (9)

4: **while** $(i \leq N_{iter})$ && $(E_{GP} > \text{threshold})$ **do**

    5:     get $F_{new}$ from solving Equation (11)

    6:     Implement experiment for $F_{new}$ and obtain $\boldsymbol{Y}(F_{new})$

    7:     $\boldsymbol{\mathcal{D}} \leftarrow \{\boldsymbol{\mathcal{D}} \cup (F_{new}, \boldsymbol{Y}(F_{new}))\}$

    8:     Estimate parameters by maximizing log-likelihood function (5) on $\boldsymbol{\mathcal{D}}$

    9:     Calculate the Fisher information matrix by solving Equations (8) and (9)

  10:     $i = i + 1$

11: **end while**

In this section, we proposed two active learning algorithms for Gaussian process model considering uncertainties. In next subsections, we will explore the initial design, evaluation criteria, and stopping criteria for these active learning algorithms.

E. *Initial Design*

The initial design defines the preliminary parameter estimation and has a significant effect on efficiency and accuracy of the model. It is critical to determine initial samples with a suitable size. The initial sample points should explore the entire space with very good space filling property. Maximin distance criterion can be used to choose a good design with space filling property. In addition, the initial samples need to provide as much information as possible. So the Latin hypercube design with good projection property is helpful. We choose the maximin Latin hypercube design to ensure the good exploration and exploitation performance for the initial samples. In addition, the size of the initial samples is very important. Under-selection of initial samples may result in insufficient parameter estimation for the proposed model, while over-selection of initial samples may reduce the efficiency and significance of the active learning algorithms. We choose the initial design by the rule of thumb of predictive modeling in composite fuselage shape control. When we run the predictive modeling in [7], we found that at least 10 training samples are needed to get a reasonable prediction performance. Therefore, we selected a maximin Latin hypercube design with 11 samples as an initial design.

F. *Stopping Criterion*

A potentially important element of interactive learning applications in general is to determine when to stop learning [1]. We take modeling for composite fuselage shape control as an example. We hope to use minimum number of training samples to get a model with satisfied prediction accuracy. It is quite complex and time-consuming to collect experimental samples in the assembly process. Therefore, it is important to know that how to recognize when the accuracy of the model can satisfy the prediction requirement, and acquiring more samples is likely increasing the flow time without much improvement on the model. Active learning provides us with the capability to accurately balance model accuracy and reducing the cost of obtaining samples.

There are several stopping criteria for active learning, such as stopping based on an intrinsic measure of stability or self-confidence within the learner [37]. In our algorithms, we check the model prediction error is consistently smaller than the engineering specifications for several continuous steps. Additional, we check that the total iteration number is smaller than a specific threshold to prevent an infinite iteration loop due to algorithm errors.

IV. CASE STUDY

We conducted a case study to demonstrate the effectiveness of the proposed active learning algorithms. Firstly, we introduced the automatic shape control of composite fuselage. Next, we discussed the parameter estimation algorithms for the predictive modeling. Three evaluation criteria have been used to evaluate the performance of active learning methods. By comparing our proposed methods with benchmark methods, we conclude that the proposed active learning methods can obtain better performance for predictive modeling. In addition, the proposed methods provide us with a decision point for stopping the collection of experimental samples.

A. *Automatic Shape Control of Composite Fuselage*

In current practice, experienced engineers adjust the shape of composite fuselage multiple times by trial-and-error, until the deviations between the real shape and the target shape are smaller than a specific engineering specification. This approach brings large uncertainties for the fuselage assembly process, and it may only reach an acceptable shape but not the optimal one. For the automatic shape control, the system is able to measure the real dimensional shape of the fuselage by a laser metrology system, then compute the optimal actuators' forces to minimize the dimensional deviations of current composite fuselage to the target one, and finally implement the shape adjustment. As shown in Fig. 1(a) [6], ten actuators are installed at the edge of the composite fuselage. These actuators can push or pull the fuselage to adjust its shape to the target shape. Du et al. investigated the optimal placement of these actuators via sparse learning [38]. The automatic shape control system can be further used for improving the dimensional quality [39] and reducing the residual stress [40] in the multistage composite structures assembly processes (e.g., automotive, aerospace). One of the most challenging tasks for automatic shape control system is to develop a predictive model with limited experimental samples. Thus, we proposed active learning strategies to maximize the information acquisition for predictive modeling and provide a stopping criterion for experiments.

In order to validate our proposed methodology economically, we developed a finite element model of the composite fuselage, with software ANSYS Composite PrepPost [6]. The finite element model, shown in Fig. 1(b), exactly mimics the fabrication process of composite fuselages, including material (carbon fiber and epoxy resin) introduction, stack-up/sub-laminates design, material orientation, geometrical setting, fixture set-up etc. Physical experiments with a fuselage from the sponsor company were conducted to calibrate and validate the accuracy of the developed computer simulation model. We applied an effective model calibration approach via sensible variable identification and adjustment [41]. After calibration, the Finite element analysis (FEA) simulation results and the physical experimental outputs are quite consistent [6, 41]. One simulation result of the total deformation is shown in Fig. 1(c). More details about the simulation and experimental setup can be found in [6, 41].

After obtaining an accurate finite element model, we generate the training and testing datasets by computer experiments. Then Gaussian process models considering uncertainties are trained to predict dimensional deviations under specific actuators' forces. More details about the performance of the Gaussian process models considering uncertainties can be found in [7].

B. *Validation Procedure of Active Learning Algorithms*

To validate the performance of active learning algorithms, we use a Gaussian process model trained by the finite element simulation datasets as an oracle model. The oracle model is to imitate the real engineering system. We generate the initial design of input by maximin Latin hypercube design. Outputs of experimental samples are generated by the oracle model with input uncertainties. For each step of active learning algorithms, we conduct parameter estimation for Gaussian process considering uncertainties by maximizing the log-likelihood function in Equation (5). The detailed procedures of active learning algorithms are shown in Algorithm 1 and Algorithm 2.

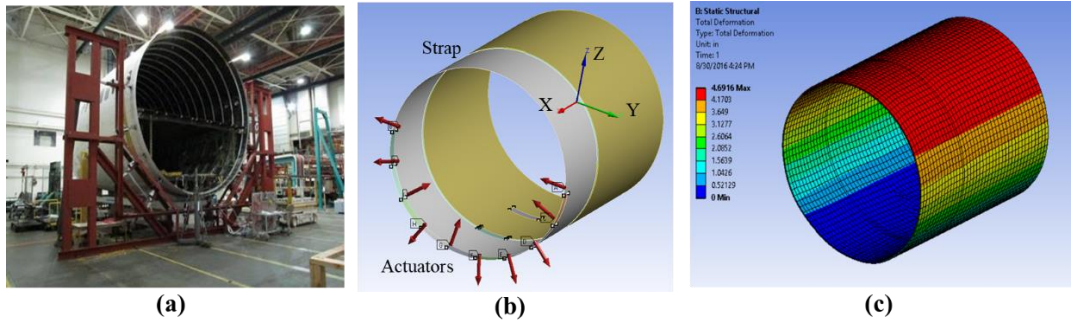

(a) Fuselage shape control [6], (b) Finite element model of shape control, (c) One simulation result
Fig. 1 Shape Control of Composite Fuselage

In the active learning algorithms, the candidate pool is generated by maximin Latin hypercube design with input bounds [-450 lbf, 450 lbf]. The size of the candidate pool is 200.

C. *Evaluation Criteria*

To evaluate the performance of active learning algorithms, we introduce three evaluation scores: mean of mean absolute deviations (mean MAD), maximum of mean absolute deviations (max MAD), and mean square error of cross-validation (cross-validation MSE). All these three evaluation scores are calculated based on the evaluation pool, which includes 200 samples explored in the whole input space. The outputs are generated by the oracle model.

The mean of mean absolute deviations (mean MAD) for each iteration can be calculated by Equation (12)

$$MAD_{mean} = \underset{j}{mean}\left\{\frac{1}{N_{eva}}\sum_{i=1}^{N_{eva}}|\widehat{\boldsymbol{Y}}_j(\boldsymbol{F}_i) - \boldsymbol{Y}_j^*(\boldsymbol{F}_i)|\right\} \quad (12)$$

where $N_{eva}$ is the size of evaluation samples. $\widehat{\boldsymbol{Y}}_j(\boldsymbol{F}_i)$ is the predictive response of the $j^{th}$ critical dimension at $\boldsymbol{F}_i$ for each iteration. It is worth mentioning that the predictive model is developed based on all samples at each iteration. $\boldsymbol{Y}_j^*(\boldsymbol{F}_i)$ is the oracle output of the $j^{th}$ critical dimension at $\boldsymbol{F}_i$.

Similarly, the max of mean absolute deviations (max MAD) can be calculated by Equation (13)

$$MAD_{max} = \underset{j}{max}\left\{\frac{1}{N_{eva}}\sum_{i=1}^{N_{eva}}|\widehat{\boldsymbol{Y}}_j(\boldsymbol{F}_i) - \boldsymbol{Y}_j^*(\boldsymbol{F}_i)|\right\} \quad (13)$$

The two evaluation scores above are based on the mean absolute deviations (MAD). MAD is an important index to check the model performance in composite fuselage shape control. We also introduce the cross-validation of mean square errors (MSE). We use leave-one-out cross-validation in which the accuracy measures are obtained by using Equation (14).

$$MSE_{cv} = \frac{1}{N_{eva}} \sum_{i=1}^{N_{eva}} \text{mean}_j \left\{ \left\| \widehat{Y}_j^{[i]}(F_i) - Y_j^*(F_i) \right\|^2 \right\} \quad (14)$$

where $\widehat{Y}_j^{[i]}(F_i)$ denotes the predicted response of the $j^{\text{th}}$ critical dimension at $F_i$, whose preditive model is trained by all the residual samples except the $i^{\text{th}}$ one.

Cross-validation is primarily a way of measuring the predictive performance of a statistical model. The procedure of calculating cross-validation MSE at each iteration includes (i) letting sample $i$ form the testing data, and $Y_j^*(F_i)$ is the oracle output of this testing sample; (ii) training the predictive model by all the residual samples except the $i^{\text{th}}$ one at this iteration, then predict the response at $F_i$, and next calculate the mean square error; (iii) repeat step (i) and (ii) for all samples at this iteration, and get the mean of obtained MSEs from step (ii). This mean is cross-validation MSE.

### D. *Comparison with Benchmark Methods*

In this subsection, we compare the performance of our proposed active learning algorithms with several benchmark methods. The first benchmark method is to obtain design samples by running design of experiment for each sample size. In this method, we do not use active learning strategy. The second benchmark method is random selection from the candidate pool, which is the most basic pool-based sampling strategy. The third benchmark method is to select the next sample which has the largest maximin distance from the current samples. This method makes full use of the space filling information in the current input variables, but not utilize the information in the response. The fourth method is expected improvement (EI) approach. The EI algorithm identifies the nature of input-output relationships, and then subsequently choose design point one at a time to maximize the expected improvement on the objective. It is widely used in sequential design, especially for computer experiments.

The evaluation criteria for these active learning methods are introduced in Section IV. C. Three evaluation scores, mean MAD, max MAD, and cross-validation MSE, are calculated as the increase of sample size. These Active learning curves are shown in Fig. 2-4. In the figures, the four benchmark methods are represented by a black dashed line (no active learning), a blue dash-dot line (random selection), a green solid line (selection based on maximin distance), and a magenta dotted line (EI algorithm). The proposed active learning algorithms are represented by a red solid line with asterisk marker (the proposed variance-based weighted active learning, VWAL algorithm) and a black solid line with plus-sign marker (the proposed D-optimal weighted active learning, DOWAL algorithm).

In Fig. 2-4, we can find that as the number of samples increases, the mean MAD becomes small for most of these methods. It makes sense because more samples tend to provide more information for training of predictive models. Without active learning, the learning curve have pretty large fluctuations. We can also find that as the number of samples increases, the MAD or MSE may become higher, e.g., number of sample from 26 to 28. One potential reason may be the sample 27 is out of the typical response surface. It means simply increasing sample size sometimes may not be a good idea. The quality of sample needs to be ensured. Further investigations to introduce an evaluation for sample quality will be conducted for the future work. In Fig. 2 and Fig. 3, among all these active learning methods, the proposed VWAL algorithm realizes the best performance and it has the smallest mean MAD and max MAD when the number of samples becomes larger than 17. But the proposed DOWAL algorithm does not have superior performance. It shows the variance-based weighted active learning algorithm is the best choice if the main objective is to realize the smallest mean or max MAD.

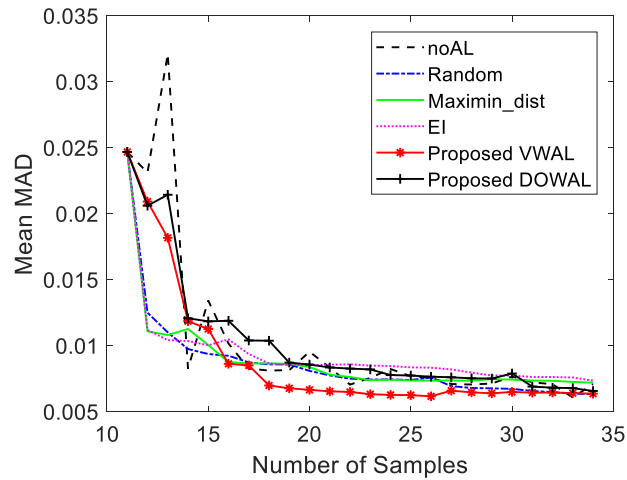

Fig. 2. Active learning curves for the mean of mean absolute deviations (MAD) of different methods

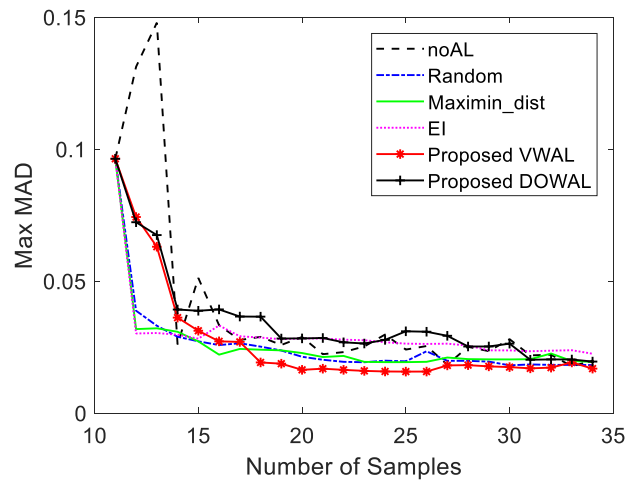

Fig. 3. Active learning curves for the maximum of mean absolute deviations (MAD) of different methods

If we take cross-validation MSE as the evaluation criterion, the proposed DOWAL algorithm realizes the best performance and has the smallest cross-validation MSE, as shown in Fig. 4. The proposed VWAL is very sensitive to the sample 27, which makes prediction become worse. It means simply increasing sample size sometimes may not be a good idea, especially when there is large nonlinearity in the response surface and the quality of samples cannot be fully ensured. From Fig. 2 to Fig. 4, we can also find the benchmark methods (e.g. random selection, selection based on maximin distance, and EI algorithm) can realize good performance. The main reason is that the candidate pool is well chosen according to maximin Latin hypercube design, and the Gaussian process model considering uncertainties can capture the main information structure and response surface within the datasets. Besides, cross-validation MSE is a widely used evaluation metric in active learning domain, while MAD is a general metric for advanced aircraft assembly applications. From the case study, we found that performance of active learning algorithms may be closely related to the evaluation criterion we selected.

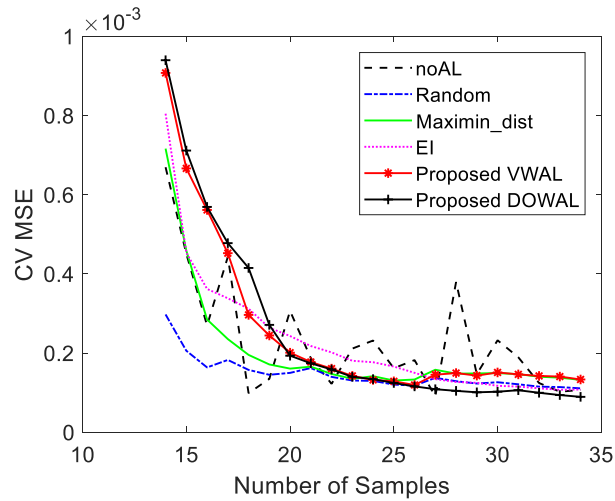

Fig. 4. Active learning curves for the cross-validation mean square errors (MSE) of different methods

As shown in Section III. F, another important element of active learning applications is knowing when to stop collecting new experimental samples. For a specific threshold from the engineering domain knowledge, e.g., mean MAD is smaller than 0.007 inches, we know if we collect 20 experimental samples, we can train a predictive model which has enough accuracy. This shows that active learning algorithms can not only improve the accuracy of predictive modeling, but also reduce the flow time for experimental sampling.

## V. Summary

To realize automatic shape control of composite fuselage, it is critical to develop an accurate predictive model. However, getting an accurate model requires many experiments by obtaining dimensional shape under different actuators' forces which are expensive and time-consuming. Thus, there is an urgent need for active learning in this application to maximize information extraction with limited experimental samples. In practice, an industrial system inevitably has numerous uncertainties, such as input uncertainties, measurement errors, modeling uncertainties, uncertainties from system parameters. Current active learning literature mainly focuses on classification problems in machine learning, and a few papers focus on regression with extrinsic uncertainties but not incorporating impacts from intrinsic input uncertainties. Therefore, there is a gap between existing active learning methods and the need to incorporate various intrinsic and extrinsic uncertainties in the advanced aircraft assembly process.

The main contribution of this paper is to propose two active learning algorithms for Gaussian process model considering uncertainties. The proposed algorithms investigate two predictive models with uncertainties: stochastic Kriging model and surrogate model considering uncertainty. We take two kinds of information measure, variance-based information and Fisher information, into consideration. Two active learning algorithms are proposed to obtain the most informative samples for Gaussian process modeling considering uncertainties. We also explored the initial design, stopping criteria for active learning algorithms. To validate the performance of the proposed algorithms, we introduced three evaluation criteria, including mean of mean absolute deviations (MAD), maximum of MAD, and cross-validation mean square errors. The proposed approaches are compared with four benchmark methods in the case study. It shows that the proposed variance-based weighted active learning algorithm can realize the best MAD performance, and the proposed D-optimal weighted active learning algorithm can realize the best cross-validation MSE performance. The proposed active learning algorithms can also be used to trigger the stopping point of experimental sampling. These strategies can be extended to other regression models with uncertainties.

ACKNOWLEDGMENT

This work was funded by the Strategic University Partnership between the Boeing Company and the Georgia Institute of Technology.

REFERENCES

[1] B. Settles, "Active learning," Synthesis Lectures on Artificial Intelligence and Machine Learning, vol. 6, no. 1, pp. 1-114, 2012.

[2] C. Q. Lam, "Sequential adaptive designs in computer experiments for response surface model fit," The Ohio State University, 2008.

[3] B. Ankenman, B. L. Nelson, and J. Staum, "Stochastic kriging for simulation metamodeling," *Operations Research*, vol. 58, no. 2, pp. 371-382, 2010.

[4] D. Cervone and N. S. Pillai, "Gaussian process regression with location errors," arXiv preprint arXiv:1506.08256, 2015.

[5] W. Wang, X. Yue, B. Haaland, and C. J. Wu, "Gaussian Process with Input Location Error and Applications to Composite Fuselage Shape Control," arXiv preprint arXiv:2002.01526, 2020.

[6] Y. Wen, X. Yue, J. H. Hunt, and J. Shi, "Feasibility analysis of composite fuselage shape control via finite element analysis," *Journal of Manufacturing Systems*, vol. 46, pp. 272-281, 2018.

[7] X. Yue, Y. Wen, J. H. Hunt, and J. Shi, "Surrogate Model-Based Control Considering Uncertainties for Composite Fuselage Assembly," *ASME Transactions, Journal of Manufacturing Science and Engineering*, vol. 140, no. 4, p. 041017, 2018.

[8] F. Olsson, "A literature survey of active machine learning in the context of natural language processing," 2009.

[9] D. D. Lewis and J. Catlett, "Heterogeneous uncertainty sampling for supervised learning," in Machine Learning Proceedings 1994: Elsevier, 1994, pp. 148-156.

[10] G. Wang, J.N. Hwang, C. Rose, and F. Wallace, "Uncertainty sampling based active learning with diversity constraint by sparse selection," In 2017 IEEE 19th International Workshop on Multimedia Signal Processing (MMSP), 2017, (pp. 1-6). IEEE.

[11] G. Wang, J.N. Hwang, C. Rose, and F. Wallace, "Uncertainty-Based Active Learning via Sparse Modeling for Image Classification," *IEEE Transactions on Image Processing*, 28(1), pp.316-329, 2018.

[12] Y. Yang, Z. Ma, F. Nie, X. Chang, and A.G. Hauptmann, "Multi-class active learning by uncertainty sampling with diversity maximization," *International Journal of Computer Vision*, 113(2), pp.113-127, 2015.

[13] D. A. Cohn, Z. Ghahramani, and M. I. Jordan, "Active learning with statistical models," *Journal of Artificial Intelligence Research*, vol. 4, pp. 129-145, 1996.

[14] M. Sugiyama, "Active learning in approximately linear regression based on conditional expectation of generalization error," *Journal of Machine Learning Research*, vol. 7, no. Jan, pp. 141-166, 2006.

[15] R. Burbidge, J. J. Rowland, and R. D. King, "Active learning for regression based on query by committee," in International Conference on Intelligent Data Engineering and Automated Learning, 2007, pp. 209-218: Springer.

[16] M. Sugiyama and N. Rubens, "Active learning with model selection in linear regression," in Proceedings of the 2008 SIAM International Conference on Data Mining, 2008, pp. 518-529: SIAM.

[17] E. Pasolli and F. Melgani, "Gaussian process regression within an active learning scheme," in Geoscience and Remote Sensing Symposium (IGARSS), 2011 IEEE International, 2011, pp. 3574-3577: IEEE.

[18] W. Cai, Y. Zhang, and J. Zhou, "Maximizing expected model change for active learning in regression," in Data Mining (ICDM), 2013 IEEE 13th International Conference on, 2013, pp. 51-60: IEEE.

[19] J. Schreiter, D. Nguyen-Tuong, M. Eberts, B. Bischoff, H. Markert, and M. Toussaint, "Safe exploration for active learning with Gaussian processes," in Joint European Conference on Machine Learning and Knowledge Discovery in Databases, 2015, pp. 133-149: Springer.

[20] T. J. Santner, B. J. Williams, and W. I. Notz, The design and analysis of computer experiments. Springer Science & Business Media, 2013.

[21] J. Sacks, W. J. Welch, T. J. Mitchell, and H. P. Wynn, "Design and analysis of computer experiments," *Statistical Science*, pp. 409-423, 1989.

[22] C. Currin, T. Mitchell, M. Morris, and D. Ylvisaker, "Bayesian prediction of deterministic functions, with applications to the design and analysis of computer experiments," *Journal of the American Statistical Association*, vol. 86, no. 416, pp. 953-963, 1991.

[23] J. Mockus, V. Tiesis, and A. Zilinskas, "Toward global optimization, volume 2, chapter bayesian methods for seeking the extremum," 1978.

[24] D. R. Jones, M. Schonlau, and W. J. Welch, "Efficient global optimization of expensive black-box functions," *Journal of Global Optimization*, vol. 13, no. 4, pp. 455-492, 1998.

[25] B. J. Williams, T. J. Santner, and W. I. Notz, "Sequential design of computer experiments to minimize integrated response functions," *Statistica Sinica*, pp. 1133-1152, 2000.

[26] B. J. Williams, T. J. Santner, W. I. Notz, and J. S. Lehman, "Sequential design of computer experiments for constrained optimization," in Statistical Modelling and Regression Structures: Springer, 2010, pp. 449-472.

[27] E. Vazquez and J. Bect, "Convergence properties of the expected improvement algorithm with fixed mean and covariance functions," *Journal of Statistical Planning and Inference*, vol. 140, no. 11, pp. 3088-3095, 2010.

[28] X. Deng, V. R. Joseph, A. Sudjianto, and C. J. Wu, "Active learning through sequential design, with applications to detection of money laundering," *Journal of the American Statistical Association*, vol. 104, no. 487, pp. 969-981, 2009.

[29] K. Crombecq, D. Gorissen, D. Deschrijver, and T. Dhaene, "A novel hybrid sequential design strategy for global surrogate modeling of computer experiments," *SIAM Journal on Scientific Computing*, vol. 33, no. 4, pp. 1948-1974, 2011.

[30] R. Jin, C.-J. Chang, and J. Shi, "Sequential measurement strategy for wafer geometric profile estimation," *IIE Transactions*, vol. 44, no. 1, pp. 1-12, 2012.

[31] H. Yan, "High dimensional data analysis for anomaly detection and quality improvement," Georgia Institute of Technology, 2017.

[32] D. Xiang, S. Gao, W. Li, X. Pu, and W. Dou, "A new nonparametric monitoring of data streams for changes in location and scale via Cucconi statistic". Journal of Nonparametric Statistics, 31(3), pp.743-760, 2019.

[33] N. Cressie, "Statistics for spatial data," Terra Nova, vol. 4, no. 5, pp. 613-617, 1992.

[34] M. L. Stein, Interpolation of spatial data: some theory for kriging. Springer Science & Business Media, 1999.

[35] V. R. Joseph and Y. Hung, "Orthogonal-maximin Latin hypercube designs," *Statistica Sinica*, pp. 171-186, 2008.

[36] K. Chaloner and I. Verdinelli, "Bayesian experimental design: A review," *Statistical Science*, pp. 273-304, 1995.

[37] F. Olsson and K. Tomanek, "An intrinsic stopping criterion for committee-based active learning," in *Proceedings of the Thirteenth Conference on Computational Natural Language Learning*, pp. 138-146, 2009.

[38] J. Du, X. Yue, J.H. Hunt, and J. Shi, "Optimal Placement of Actuators Via Sparse Learning for Composite Fuselage Shape Control". *Journal of Manufacturing Science and Engineering*, 141(10), 2019.

[39] X. Yue, and J. Shi, "Surrogate model–based optimal feed-forward control for dimensional-variation reduction in composite parts' assembly processes". *Journal of Quality Technology*, 50(3), pp.279-289, 2018.

[40] Y. Wen, X. Yue, J. H. Hunt, and J. Shi, "Virtual assembly and residual stress analysis for the composite fuselage assembly process". *Journal of Manufacturing Systems*, 52, pp.55-62, 2019.

[41] Y. Wang, X. Yue, R. Tuo, J. H. Hunt, and J. Shi, "Effective Model Calibration via Sensible Variable Identification and Adjustment, with application to Composite Fuselage Simulation," arXiv preprint arXiv:1912.12569, 2019.